# Phase Transitions of Plan Modification in Conformant Planning

**Junping Zhou**[1,2], **Minghao Yin**[1,2,*]

[1] *School of Computer Science, Northeast Normal University, Changchun 130012, China*

[2] *Key Laboratory of Symbolic Computation and Knowledge Engineering for the Ministry of Education, Jilin University, Changchun 130012, China*

Correspondence should be addressed to Minghao Yin, ymh@nenu.edu.cn

We explore phase transitions of plan modification, which mainly focus on the conformant planning problems. By analyzing features of plan modification in conformant planning problems, quantitative results are obtained. If the number of operators is less than $\alpha_{ub}$, almost all conformant planning problems can't be solved with plan modification. If the number of operators is more than $\alpha_{lb}$, almost all conformant planning problems can be solved with plan modification. The results of the experiments also show that there exists an experimental threshold $\alpha_c$ of density (ratio of number of operators to number of propositions), which separates the region where almost all conformant planning problems can't be solved with plan modification from the region where almost all conformant planning problems can be solved with plan modification.

## 1. Introduction

Artificial Intelligence (AI) planning involves generating a series of operators leading from the initial world description to the goal world description[1]. Whenever a system needs to take into account operators and goals, AI planning is an important component. Therefore, AI planning techniques have been widely used in many engineering problems. For example, classical planning has successfully applied to autonomous controller for NASA's *Deep Space O*ne spacecraft, scheduled to be launched in late 1998[2]. Off-the-shelf conformant planning tools are exploited to solve message-based Web Service Composition in a general form that involves powerful ontologies[3].

As an extension of classical planning, conformant planning is the problem of finding a sequence of operators for achieving a goal in the presence of uncertainty in the operators or initial state[4,5]. Since conformant planning is computationally harder than classical planning, researchers explore approaches of solving the problem and plan modification is one of them. Plan modification is formalized as a process of removing inconsistencies in the validation structure of a plan when it is being reused in a new or changed planning situation[6]. Without generating a plan from scratch, which is actually a process of resource and time consuming, plan modification has long been recognized as a valuable tool for the improvement of efficiency in planning systems. Recently, Nebel and Koehler in [7] discuss the relationship between plan modification and plan generation pointing out that sometimes modifying a plan is not computationally as easy as planning from scratch. In consequence, some most interesting questions are put forward. Under which circumstances it is worthwhile to adopt plan modification and under which circumstances it isn't worthwhile to adopt plan modification. Whether there exists a rapid transition between almost all plan problems are solved with plan modification and almost all plan problems aren't solved with plan modification.

In this paper, we conduct research on phase transitions further up the polynomial hierarchy in problems like plan modification in conformant planning problems, whose computational complexity is EXPSPACE-complete. By analyzing features of plan modification in conformant planning problems, lower and upper bounds are obtained. When $o$ is less than $\alpha_{ub}$, almost all conformant planning problems can't be solved with plan modification. And when $o$ is more than $\alpha_{lb}$, almost all conformant planning problems can be solved with plan modification. The results of the experiments show that there exists an experimental threshold $\alpha_c$ of density (ratio of number of operators to number of propositions), which separates the region where almost all conformant planning problems can't be solved with plan modification from the region where almost all conformant planning problems can be solved with plan modification.

## 2. Related Work

Regarding conformant planning, Smith and Weld present a conformant planning solver—CGP which is a Graphplan-based planner[5]. Bertoli and Cimatti tackle conformant planning via model checking which relies on symbolic techniques such as Binary Decision Diagrams to compactly represent and efficiently analyze the planning domain[8]. Geffner et al address an integrated software tool for modeling, analyzing and solving conformant planning problems, contingent planning problems, etc[9]. Hoffmann and Brafman give a new representation of search space and extend the classical heuristic planning system FF to the conformant setting[10]. Palacios and Geffner propose CF2CS, which has won the conformant track of IPC-5. They adopt a polynomial scheme for mapping conformant problems with deterministic operators into classical problems[11]. Dunbo Cai et al design a conformant planning heuristic function based on a classical one in a way that pursues plan reuse between world states in the same belief state[12].

Regarding plan modification, Kambhampati et al present an approach for flexible reuse of old plans in the presence of a generative planner[13]. Later, he addresses a theory of plan modification applicable to hierarchical nonlinear planning[14]. Hendler et al study the validation-structure-based theory of plan modification and reuse[6]. Chuan Wu et al extend the plan modification to medical science[15]. Koehler presents a domain-independent approach to flexible plan reuse based on a deductive framework[16]. Verfaillie and Schiex explore the solution reuse in dynamic constraint satisfaction problems[17]. Nebel and Koehler in [5] discuss the relationship between plan modification and plan generation and point out that sometimes modifying a plan is not computationally as easy as planning from scratch.

Regarding phase transaction in intractable problems, Kirkpatrick and Selman in [18] indicate that the propositional satisfiability problem, or $k$-SAT problem for short, takes place the phase transition phenomena. If $\alpha < \alpha_c$ ($\alpha$ and $\alpha_c$ are the ratios of number of clauses to number of variables), $k$-SAT problem is satisfiable with probability tending to 1. If $\alpha > \alpha_c$, $k$-SAT problem is satisfiable with probability tending to 0. Gent and Walsh point out that there is a rapid transition between soluble and insoluble instances in the traveling salesman problem and hard instances of the traveling salesman problem are associated with this transition [19]. Ke Xu et al propose a new type of random CSP model, with which they prove that the phase transitions exist in CSP[20]. Coppersmith et al study the Max-SAT phase transitions[21]. Gent and Walsh show that phase transition behavior similar to SAT problems occurs in QBF problems[22]. Bailey et al investigate the existence of phase transitions for a family of PP-complete problems—#SAT problems[23].

## 3. Random Conformant Planning Models

In this section, we mainly discuss random conformant planning models to carry out an investigation on phase transitions of plan modification in conformant planning problems. From now on, unless otherwise stated, $n$ denotes the number of propositions; $o$ denotes the number of operators; $r$ and $c$ respectively denote the numbers of preconditions and postconditions within an operator; $g$ denotes the number of goal conditions; $k$ denotes the number of initial states.

**Definition 3.1. (Conformant Planning Problem)** A conformant planning problem can be regarded as a 4-tuple $P = \langle S, A, I, G \rangle$, where:
- $S$ is a finite set of states and the subset of $S$ is called belief state;
- $A$ is a finite set of operators. An operator is a pair <*pre, eff*> respectively called precondition and effect. The effect is a triple <*cond, del, add*>, where *cond* is the effect conditions, *del* is the delete effects and *add* is the add effects;
- $I \subseteq S$ is a set of initial states over $S$, which is also called initial belief state;
- $G \neq \varnothing$ is the goal, which is actually a partial state composed of goal conditions.

Conformant planning problem is the task of generating plans when the initial state is partially known and the operators can have non-deterministic effects. In particular, in this paper we only focus on the conformant planning problems that the initial state is partially known and the operators have deterministic effects. The reason is that nearly all researches on conformant planning problems are in the assumption that operators are all deterministic and all uncertainty lies in the initial situation. Moreover the assumption can't reduce the computational complexity. In [24], Rintanen has proved that the computational complexity of conformant planning problems is EXPSPACE-complete. This also holds for problems with deterministic operators.

**Definition 3.2. (Conformant Plan)** A conformant plan is a sequence of operators $\pi = <a_0, a_1, \ldots, a_{n-1}>$ iff $G \subseteq \pi(I)$.

**Definition 3.3. (Variable Model)** Let $n$ be the number of propositions; $o$ be the number of operators; $r$ and $c$ be the numbers of preconditions and postconditions within an operator respectively; $g$ be the number of goal conditions; $k$ be the number of initial states. A variable model $M_v$ is a 4-tuple $\langle S, A, I, G \rangle$, where:
- $S$ is a finite set of states and every state is made up of some propositions;
- $A$ is a finite set of operators with $o$ operators. For $\forall a_1, a_2 \in A$, $Num(pre(a_1)) \neq Num(pre(a_2))$ and $Num(eff(a_1)) \neq Num(eff(a_2))$ are allowed, where $Num(pre(a_1))$ and $Num(pre(a_2))$ are the numbers of preconditions of $a_1$ and $a_2$ respectively, and $Num(eff(a_1))$ and $Num(eff(a_2))$ are the numbers of postconditons of $a_1$ and $a_2$ respectively;
- $I \subseteq S$ is a set of initial states over $S$, which is also called the initial belief state;
- $G \neq \varnothing$ is the goal, which is actually a partial state composed of goal conditions.

**Definition 3.4. (Fixed Model)** Let $n$ be the number of propositions; $o$ be the number of operators; $r$ and $c$ respectively be the numbers of preconditions and postconditions within an operator; $g$ be the number of goal conditions; $k$ be the number of initial states. A fixed model $M_f$ is a 4-tuple $\langle S, A, I, G \rangle$, where:
- $S$ is a finite set of states and every state is made up of some propositions;
- $A$ is a finite set of operators with $o$ operators. For $\forall a_1, a_2 \in A$, there only exists $Num(pre(a_1)) = Num(pre(a_2))$ and $Num(eff(a_1)) = Num(eff(a_2))$, where $Num(pre(a_1))$ and $Num(pre(a_2))$ are the numbers of preconditions of $a_1$ and $a_2$ respectively, $Num(eff(a_1))$ and $Num(eff(a_2))$ are the numbers of postconditons of $a_1$ and $a_2$ respectively;
- $I \subseteq S$ is a set of states over $S$, which is also called the initial belief state;
- $G \neq \varnothing$ is the goal, which is actually a partial state composed of goal conditions.

In fact, the difference between the variable model and the fixed model is whether each operator has the same number of preconditions and postconditons respectively. According to the definitions of variable model and fixed model, the conformant planning instances generated in this paper comply with two assumptions. The first one is that each precondition of an operator is selected independently of other precondition and postcondition. The same rule also holds for the postconditons. The second one is that each operator has fixed numbers of preconditions and postconditions.

## 4. Plan Modification in Conformant Planning

Plan modification is to construct a plan for a new planning problem by modifying the existing plans. The following definition addresses the plan modification problem in conformant planning.

**Definition 4.1. (Plan Modification Problem in Conformant Planning)** A plan modification problem in conformant planning is given as follows:
Given a conformant planning problem $P' = \langle S', A, I', G' \rangle$ and a conformant plan $CP$ that solves the conformant planning problem $P = \langle S, A, I, G \rangle$, a conformant plan $CP'$ is generated for $P'$ by minimally modifying $CP$.

In this paper, we consider a simplified version of plan modification problem in conformant planning. Given a conformant planning problem $P' = \langle S', A, I', G' \rangle$ and a conformant plan $CP$ that solves the conformant planning problem $P = \langle S, A, I, G \rangle$, where $S$ and $S'$ are made up of the same set of propositions, the difference between the old and new problem situations is only one proposition, specifically:
- $G$ differs by one proposition from $G'$, or
- $I$ differs by one proposition from $I'$. In view of $I$ and $I'$ being both belief states, the only one proposition relates to all states in $I$ and $I'$.

A conformant plan $CP'$ is generated for $P'$ by minimally modifying $CP$.

In the first case, the conformant plan $CP$ doesn't need to be modified when the number of propositions in $G$ is more than $G'$. However, when the number of propositions in $G'$ is more than $G$, the plan modification problem in conformant planning can be viewed as a conformant planning problem with *n* goal conditions, where *n*-1 goal conditions are true of in the initial belief state.

In the second case, the conformant plan *CP* doesn't need to be modified when the number of propositions in *I'* is more than *I*. However, when the number of propositions in *I* is more than *I'*, the plan modification problem in conformant planning can be viewed as a conformant planning problem with *n* goal conditions, where *n*-1 goal conditions are true of in the initial belief state.

The reason why we consider this simplified version is that when the old and new problem situations differ by some propositions, the plan modification problem can be regarded as many this simplified version of plan modifications. In other words, in order to solve the new planning problem, we can construct a sequence of planning problems, two of which differ by one proposition, from the old planning problem situation to the new planning problem situation. Therefore, with the purpose of analyzing the phase transition of plan modification in conformant planning problems, we only need to discuss the lower bound and the upper bound on conformant planning problems with *n* goal conditions, where *n*-1 goal conditions are true of in the initial belief state.

## 5. Lower and Upper Bounds

The lower and upper bounds are used to guide under which circumstances it is worthwhile to adopt plan modification and under which circumstances it isn't worthwhile to adopt plan modification. Owing to the existence of lower bound and upper bound, the region is separated where almost all conformant planning problems can be solved with plan modification from where almost all conformant planning problems can't be solved with plan modification. In the following, the analysis of lower and upper bounds is based on the assumption that readers are familiar with the notions about the conformant planning problems, such as the exclusive operators, consistent operators, etc.

### 5.1. *The Fundamental Analysis of Variable Model and Fixed Model*

This section presents the distributions of conformant planning instances under the variable model and the fixed model. Because of the assumption that each precondition(or postconditon) of an operator is selected independently of other preconditions and postconditions, the distribution of random conformant planning instances under the variable model are given as follows:

Given *n*, *o*, *r*, *c*, and *g*, for a random proposition *p* and operator *a*, *p* is a precondition of the operator with probability $r/(2n)$; alternatively $\neg p$ is a precondition with probability $r/(2n)$. And $c/(2n)$ is the probability for postconditions.

For random conformant planning instances under the fixed model, lemma 5.1 presents the distributions of these instances.

**Lemma 5.1.** Let *n* be the number of propositions; *o* be the number of operators; *r* and *c* respectively be the numbers of preconditions and postconditions within an operator; *g* be the number of goal conditions; *k* be the number of initial states. Given a random instance under the fixed model, then

$$f(j, n, w) = \frac{n-w}{n} f(j\text{-}1, n\text{-}1, w) + \frac{w}{2n} f(j\text{-}1, n\text{-}1, w\text{-}1). \tag{5.1}$$

where $f(j, n, w)$ is the probability that *j* conditions can be randomly generated from *n* propositions so that they are consistent with some particular set of *w* conditions.

*Proof*: The probability $f(j, n, w)$ can be divided into two probabilities. One is the probability $p_1$ that a condition is randomly generated from the *n* propositions and it is neither identical to nor the negation of one of the *w* conditions; the other probability $p_2$ is that a condition is generated from the *n* propositions and is identical to one of the *w* conditions.

In the first case, the probability $p_1$ that a condition is randomly generated from the *n* propositions and it is neither identical to nor the negation of one of the *w* conditions is as follows.

$$p_1 = \frac{n-w}{n} f(j\text{-}1, n\text{-}1, w). \tag{5.2}$$

In the second case, the probability $p_2$ that a condition is generated from the *n* propositions is identical to one of the $w$ conditions is as follows.

$$P_2 = \frac{w}{2n} f(j-1, n-1, w-1). \tag{5.3}$$

Thus, the probability $f(j, n, w)$ is

$$f(j, n, w) = p_1 + P_2 = \frac{n-w}{n} f(j-1, n-1, w) + \frac{w}{2n} f(j-1, n-1, w-1). \tag{5.4}$$

Therefore, Equality (5.1) is satisfied. □

In the base cases, the probability is 1 if there are no conditions to select or no conditions to be consistent with, i.e. $f(0, n, w) = 1$, $f(j, n, 0) = 1$; the probability is $f(j, n, n) = 2^{-j}$ if each condition to be consistent with must have a particular sign; the probability is $f(n, n, w) = 2^{-w}$ if each condition to be consistent with must be selected.

In addition, in the rest of the paper, two inequalities are also used to analyze the lower bound and upper bound and the detail proof process can be seen in [25].

$$e^{-x/(1-x)} \leq 1-x \quad \text{for } 0 \leq x \leq 1. \tag{5.5}$$

$$1-x \leq e^{-x} \quad \text{for } x \geq 0. \tag{5.6}$$

In fact, with the fundamental analysis of variable model and fixed model we obtain the following upper bound and lower bound.

## 5.2 Upper Bound

In this section, we discuss the upper bound of random conformant planning instances in which there are $n$ goal conditions, where $n-1$ goal conditions are true of the initial belief state. The analysis is carried out under the variable model and the fixed model respectively. We thus obtain the following theorems:

**Theorem 5.2.** For random conformant planning instances under the variable model in which there are $n$ goal conditions, where $n-1$ goal conditions are true of the initial belief state, *if*:

$$o \leq \alpha_{ub}, \text{ where } \alpha_{ub} = \log^{(1-\sigma)}_{(1-c/2n)}. \tag{5.7}$$

where $n$ is the number of propositions; $o$ is the number of operators; $r$ and $c$ respectively are the numbers of preconditions and postconditions within an operator; $k$ is the number of initial states. Then, for at least 1-$\sigma$ of the instances, where $\sigma$ is a constant, no operators solve the instance in one step.

*Proof*: If the only unachieved goal is not a postcondition of any operator, then there will be no operators that can solve the instance in one step. Thus, we discuss the probability that a goal is not a postcondition of any operator.

$c/2n$      probability that the goal is a postcondition of the operator
$1-c/2n$      probability that the goal is not a postcondition of the operator
$(1-c/2n)^o$      probability that the goal is not a postcondition of any operator
$1-(1-c/2n)^o$      probability that the goal is a postcondition of some operator

By the Equality (5.7) we have:

$$o \leq \ln(1-\sigma)/\ln(1-c/2n). \tag{5.8}$$

This is equivalent to:

$$\ln(1-\sigma) \leq o \ln(1-c/2n). \tag{5.9}$$

and:

$$1-\sigma \leq (1-c/2n)^o. \tag{5.10}$$

finally:

$$1-(1-c/2n)^o \leq \sigma. \tag{5.11}$$

Therefore, if the number of operators satisfies Equality (5.7), for at least $1-\sigma$ of the instances, no operators solve the instance in one step. □

**Theorem 5.3.** For random conformant planning instances under the fixed model in which there are $n$ goal conditions, where $n$-1 goal conditions are true of the initial belief state, *if*:

$$o \leq \alpha_{ub}, \text{ where } \alpha_{ub} = \log_{(1-c/2n)}^{(1-\sigma)}. \tag{5.12}$$

where $n$ is the number of propositions; $o$ is the number of operators; $r$ and $c$ respectively are the numbers of preconditions and postconditions within an operator; $k$ is the number of initial states. Then, for at least $1-\sigma$ of the instances, where $\sigma$ is a constant, no operators solve the instance in one step.

*Proof*: The process presented in the previous proof is also hold for the fixed model. Because in the fixed model, $1-(1-c/2n)^o$ is also the probability that a particular goal is a postcondition of some operator. Therefore, the same inequality holds for the fixed model. □

From Theorem 5.2 and Theorem 5.3, we find that if the number of operators is less than the upper bound $\alpha_{ub}$, no operators can solve the conformant planning problem in one step. Hence, it isn't worthwhile to solve the random conformant planning instances with plan modification.

## 5.3 *Lower Bound*

The upper bound shows under which circumstances plan modification doesn't work. In this section, we give the lower bound in order to point out in which cases conformant planning problems can be solved with plan modification.

**Theorem 5.4** For random conformant planning instances under the variable model in which there are $n$ goal conditions, where $n$-1 goal conditions are true of the initial belief state, *if*:

$$o \geq \alpha_{lb}, \text{ where } \alpha_{lb} = e^{rk} \ e^c \ (2n/c) \ (\ln 1/\sigma). \tag{5.13}$$

where $n$ is the number of propositions; $o$ is the number of operators; $r$ and $c$ respectively are the numbers of preconditions and postconditions within an operator; $k$ is the number of initial states. Then, for at least $1-\sigma$ of the instances, where $\sigma$ is a constant, some operator solves the instance in one step.

*Proof*: Let $p$ be the probability that a random operator can solve a random conformant planning instance under the variable model.

Firstly, the probability that the preconditions of a random operator are consistent with the $k$ initial states is considered.

$(1-r/2n)^n$     probability that a state satisfies the preconditions of an operator
$(1-r/2n)^{nk}$     probability that $k$ initial states satisfy the preconditions of an operator, i.e. the probability that the preconditions of a random operator are consistent with the $k$ initial states

Secondly, the probability that the postconditions of an operator are consistent with the $n$-1 achieved goals is presented.

$c/2n$     probability that a goal is a postcondition of the operator
$1-c/2n$     probability that a goal is not a postcondition of the operator
$(1-c/2n)^{n-1}$     probability that $n$-1 goals are not a postcondition of the operator, i.e. the probability that the postconditions of an operator are consistent with the $n$-1 achieved goals

Thirdly, the probability that the goal is achieved by a postcondition is given.

$c/2n$ probability that a goal is a postcondition of an operator

Then the probability $p$ that a random operator can solve a random instance under the variable model can be written as follows:

$$p = (1 - r/2n)^{nk} (1 - c/2n)^{n-1} (c/2n). \tag{5.14}$$

Let $x = r/2n$, from Inequality (5.5), we have:

$$(1 - r/2n)^{nk} \geq e^{-rnk/(2n-r)} \geq e^{-rk}. \tag{5.15}$$

Let $x = c/2n$, from Inequality (5.5), we have:

$$(1 - c/2n)^{n-1} \geq e^{-c(n-1)/(2n-c)} \geq e^{-c}. \tag{5.16}$$

Thus by Inequality (5.15) and (5.16) we get:

$$p \geq e^{-rk} e^{-c} (c/2n). \tag{5.17}$$

$1 - p$ is the probability that an operator can't solve a random instance, and $(1-p)^o$ is the probability that $o$ operators can't solve a random instance.

If $o$ satisfies the Inequality (5.13), from Inequality (5.6), we have:

$$(1 - p)^o \leq e^{-po} \leq e^{-\ln 1/\sigma} = \sigma. \tag{5.18}$$

Therefore there will be at least $1 - \sigma$ of the instances under the variable model, some operator solves the instance in one step. □

**Theorem 5.5.** For random conformant planning instances under the fixed model in which there are $n$ goal conditions, where $n-1$ goal conditions are true of the initial belief state, *if*:

$$o \geq \alpha_{lb}, \text{ where } \alpha_{lb} = e^{rk} e^{c} (2n/c) (\ln 1/\sigma). \tag{5.19}$$

where $n$ is the number of propositions; $o$ is the number of operators; $r$ and $c$ respectively are the numbers of preconditions and postconditions within an operator; $k$ is the number of initial states. Then, for at least $1 - \sigma$ of the instances, where $\sigma$ is a constant, some operator solves the instance in one step.

*Proof*: Let $p$ be the probability that a random operator can solve a random instance under the fixed model.

The probability that the preconditions of a random operator are consistent with the $k$ initial states is $2^{-rk}$ and

$$2^{-rk} \geq e^{-rk}. \tag{5.20}$$

The probability that the postconditions of an operator are consistent with the $n-1$ achieved goals is $f(c-1, n-1, n-1)$ and

$$f(c-1, n-1, n-1) \geq 2^{-c+1} \geq e^{-c}. \tag{5.21}$$

where $f$ is defined by Equation (5.1).

The probability that a goal is a postcondition of the operator is $c/2n$.

Thus, the probability $p$ that a random operator can solve a random instance under the fixed model is the following:

$$p = 2^{-rk} f(c-1, n-1, n-1) (c/2n). \tag{5.22}$$

By Inequality (5.20) and (5.21) we get:

$$p \geq e^{-rk} e^{-c} (c/2n). \tag{5.23}$$

$1-p$ is the probability that an operator can't solve a random instance, and $(1-p)^o$ is the probability that $o$ operators can't solve a random instance.
If $o$ satisfies the Inequality (5.19), we have

$$(1-p)^o \leq e^{-po} \leq e^{-\ln 1/\sigma} = \sigma. \tag{5.24}$$

Therefore, there will be at least $1-\sigma$ of the instances under the fixed model, some operator solves the instance in one step. □

Theorem 5.4 and Theorem 5.5 address under which circumstances plan modification works. Summary, the above theorems show under which circumstances plan modification works and these guide whether the conformant planning problems can be solved with plan modification or not. Owing the existence of the lower and upper bounds, the region is separated where almost all conformant planning problems can be solved with plan modification from where almost all conformant planning problems can't be solved with plan modification.

## 6 Experimental Results

In order to prove the phase transitions do exist in conformant plan modification, we have experiments on conformant planning instances under the improved variable model. Since the plan modification problem in conformant planning can be viewed as a conformant planning problem with $n$ goal conditions, where $n$-1 goal conditions are true of in the initial belief state, each generated instance has three preconditions and two postconditions within an operator; $2^m$ ($m$ is a constant) initial states; one goal condition. In this experiments, we test a large collection of conformant planning instances with 10, 15, 20, 30, 40, 50 propositions for each density $\alpha$ (ratio of number of operators to number of propositions). All experiments are run on a Linux 2.0*2 GHz with 1GB RAM.

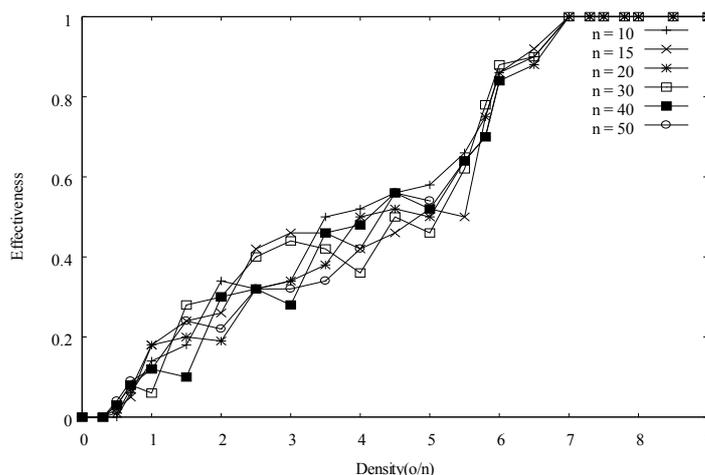

**Fig. 1:** The results for the phase transition of plan modification in conformant planning problems

The results for the phase transitions of plan modification in conformant planning problems are presented in Fig.1. From Fig.1 we find out that the problem instances turn from almost certainly insoluble to almost certainly soluble when they execute some operator with the increasing densities. The reason is that the probability of the goal condition satisfied by some operator grows when the number of propositions and the number of goal conditions are unchangeable and the densities raise. The phase transitions curve depicted in Fig. 1 matches the above theoretic analysis. The experiments show that there indeed exists an experimental density $\alpha_c$, which separates the region where almost all conformant planning problems can be solved with plan modification from the region where almost all conformant planning problems can't be solved with plan modification.

# 7 Conclusions

In this paper, we explore the phase transitions of plan modification in conformant planning problems. By analyzing features of the conformant plan modification, quantitative results are obtained. If the number of operators is less than $\alpha_{ub}$, almost all conformant planning problems can't be solved with plan modification. If the number of operators is more than $\alpha_{lb}$, almost all conformant planning problems can be solved with plan modification. The results of the experiments show that there exists an experimental threshold $\alpha_c$ of density, which separates the region where almost all conformant planning problems can't be solved with plan modification from the region where almost all conformant planning problems can be solved with plan modification. In the future, we would carry on the research about the plan modification in narrowing the gap between the bounds.

# Acknowledgment

This project was granted by the National Natural Science Foundation of China under Grant Nos.60496321, 60803102, 60673099, 60873146; the Specialized Research Fund for the Doctoral Program of Higher Education of China under Grant No.20050183065; National High Technology Research and Development Program of China under Grant Nos. 2007AAO4Z114; 2009AA02Z307.